\documentclass{article}
\usepackage{spconf,amsmath,graphicx}
\usepackage[colorlinks=true, citecolor=green]{hyperref}
\usepackage{indentfirst}
\usepackage{booktabs}
\usepackage{amssymb}
\usepackage{xspace}
\usepackage{xcolor,colortbl}
\usepackage{pifont}
\usepackage{comment}
\usepackage{bm}

\usepackage[capitalize]{cleveref}
\crefname{section}{Sec.}{Secs.}
\Crefname{section}{Section}{Sections}
\crefname{table}{Tab.}{Tabs.}
\Crefname{table}{Table}{Tables}

\newcommand*{\method}{EventEgoHands\xspace}
\newcommand*{\dataset}{N-HOT3D\xspace}

\definecolor{lightgray}{rgb}{0.88, 0.88, 0.88}

\title{EventEgoHands: Event-based Egocentric 3D Hand Mesh Reconstruction}

\name{Ryosei Hara$^{\dagger}$ \qquad Wataru Ikeda$^{\dagger}$ \qquad Masashi Hatano$^{\dagger}$ \qquad Mariko Isogawa$^{\dagger \ddag}$\thanks{supplementary materials: \href{https://sigport.org/documents/supplementary-material-eventegohands-event-based-egocentric-3d-hand-mesh-reconstruction}{Link}}}
\address{$^{\dagger}$Keio University, $^{\ddag}$JST Presto}

\begin{document}
%
\maketitle
%

\begin{abstract} 
\hspace{1em}
Reconstructing 3D hand mesh is challenging but an important task for human-computer interaction and AR/VR applications.
In particular, RGB and/or depth cameras have been widely used in this task.
However, methods using these conventional cameras face challenges in low-light environments and during motion blur. 
Thus, to address these limitations, event cameras have been attracting attention in recent years for their high dynamic range and high temporal resolution. 
Despite their advantages, event cameras are sensitive to background noise or camera motion, which has limited existing studies to static backgrounds and fixed cameras.
In this study, we propose \method, a novel method for event-based 3D hand mesh reconstruction in an egocentric view. 
Our approach introduces a Hand Segmentation Module that extracts hand regions, effectively mitigating the influence of dynamic background events.
We evaluated our approach and demonstrated its effectiveness on the \dataset dataset, improving MPJPE by approximately more than $4.5~\mathrm{cm}$~(43\%).
\end{abstract}
%

\section{Introduction}
\label{sec:intro}

\begin{figure}[t]
\centerline{\includegraphics[width=\linewidth]{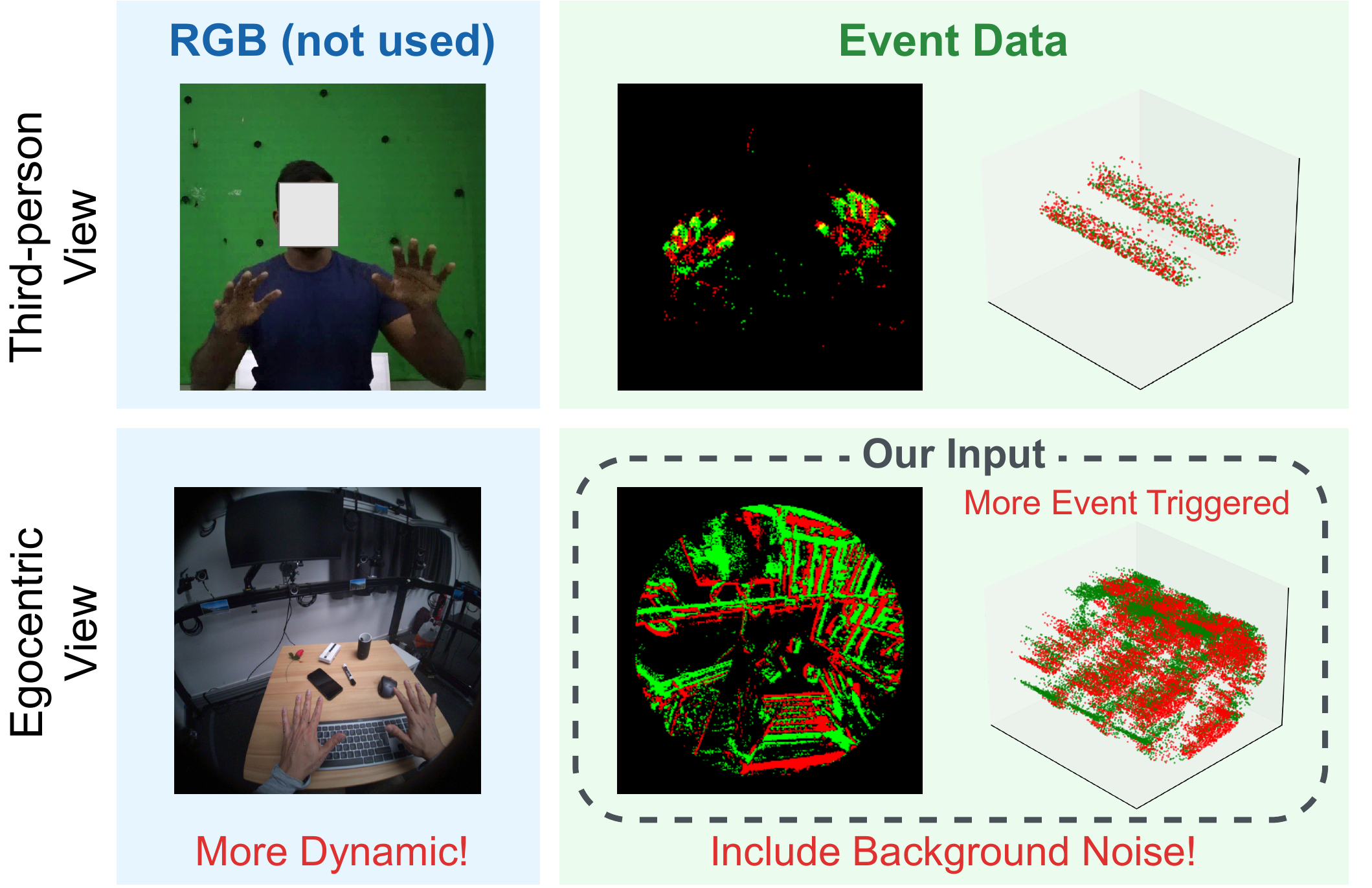}}
\caption{
\textbf{Egocentric event camera problem}.
A fixed third-person view is limited to specific scenarios, while an egocentric view offers greater flexibility and mobility. However, in an egocentric event camera setup, the camera wearer's movements can generate numerous background events, making it challenging to accurately recognize the hands.
}
\label{fig:teaser}
\end{figure}

3D human hand mesh reconstruction has many applications, and many methods have been proposed up to now~\cite{everyday_ego, hamer, AssemblyHands, handdagt, EgoRGBD}.
In particular, methods using egocentric cameras~\cite{everyday_ego, AssemblyHands, EgoRGBD} are essential for applications such as AR/VR, and robotics.

However, all of these existing methods rely on RGB(D)-based approaches and they have two main challenges.
First, it is difficult for the methods to recognize hands in low-light environments.
In the egocentric video, the brightness of the environment frequently changes more than in the third-person perspective videos due to the movement of the person wearing the camera.
Second, it is vulnerable to motion blur. 
Rapid hand movements and the movement of the camera wearer result in subject blurring, which reduces the accuracy of the estimation.

The event-based camera~(hereafter, event camera), which captures luminance changes asynchronously, offers a promising solution due to their high dynamic range and high temporal resolution~\cite{eventsurvey}.
Their high dynamic range provides high resistance to changes in brightness, making them effective not only in dark environments, but also in extremely bright conditions where overexposure can occur.
Additionally, their high temporal resolution allows for the precise capture of rapid hand movements.
They have been applied to various tasks, such as optical flow estimation~\cite{optical_flow_1}, semantic segmentation~\cite{semantic_segmentation_1}, and human pose estimation~\cite{EventEgo3D, EventHPE}.

3D hand mesh reconstruction methods using an event camera~\cite{EventHands, evhandpose, ev2hands} have also been proposed.
Although these methods leverage the advantages of event cameras to achieve robust and fast estimation, even in low-light conditions, they assume a limited scenario: a fixed third-person camera setup and static backgrounds, which significantly limit practicality.

Therefore, we tackled the novel task of egocentric event-based 3D hand mesh reconstruction.
The challenges of egocentric event camera-based method are summarized in \cref{fig:teaser}.
In the egocentric scenario, the camera is no longer static because the camera-wearer moves. 
This camera motion generates a significant number of irrelevant background events to hand mesh reconstruction, making the task more challenging. 
In addition, the computational cost is high for processing a substantial number of events.

To address these challenges, we propose {\bf{\method}}, a framework that effectively reconstructs the human hand shape using with egocentric event data.
To leverage only the events triggered around the hand region by eliminating irrelevant background events, we also propose a Hand Segmentation Module. This mask is utilized to extract event data effectively while also contributing to a computational cost reduction by decreasing the amount of data to be processed.

In addition, no dataset is available specifically for event-based hand mesh reconstruction from an egocentric perspective. 
Therefore, we created \dataset, a large-scale synthetic dataset with $447{,}704$ samples. \dataset is generated by simulation with event simulator v2e~\cite{v2e}, based on HOT3D~\cite{hot3d}, a conventional camera-based egocentric dataset for 3D hand and object tracking.

Our technical contributions are as follows:
\begin{itemize}
    \item The first approach for egocentric event-based 3D hand mesh reconstruction.
    \item We proposed Hand Segmentation Module mitigates the influence of background events caused by egocentric camera motion. This module estimates a hand mask and leverages the event data within the hand mask for 3D hand mesh reconstruction.
    \item We created \dataset, the first synthetic event-based hand dataset designed specifically for egocentric perspectives.  The dataset contains a total of 447K samples.
    \item Through both quantitative and qualitative evaluations, we demonstrated the effectiveness of \method, improving R-AUC by 0.2 points~(79\%), MPJPE by over $4.5~\mathrm{cm}$~(43\%) and MPVPE by more than $2.2~\mathrm{cm}$~(34\%) compared to previous methods.
\end{itemize}

\section{RELATED WORKS}
\label{sec:related}

\subsection{3D Hand Mesh Reconstruction}
\label{ssec:rgbhand}
3D hand mesh reconstruction has seen significant advancements, most of which rely on RGB~\cite{everyday_ego, hamer, AssemblyHands} or depth~\cite{handdagt, EgoRGBD} sensors.
Parametric models, such as MANO~\cite{MANO}, are often used to represent hand meshes for reconstruction. 
In this study, we also utilized MANO parameters provided by the HOT3D~\cite{hot3d} dataset.
These standard approaches have not worked well in dark conditions and motion blur caused by the quick movement of hand and camera.
As a solution to these issues, this study explores the method that utilizes an event camera.

\subsection{Event-based 3D Hand Mesh Reconstruction}
\label{ssec:eventhand}
Recently, event cameras have gained popularity~\cite{eventsurvey} due to their high dynamic range and high temporal resolution, addressing challenges such as low-light conditions and motion blur that conventional sensors (\eg, RGB or depth) often struggle with.
Event cameras generate asynchronous event streams by recording changes in the intensity of each pixel. 

Several studies~\cite{EventHands, ev2hands, evhandpose} address the 3D hand mesh reconstruction task using an event camera.
Rudnev~\etal~\cite{EventHands} have developed a lightweight framework for fast hand motion estimation. They utilized a frame-based 2D representation called Locally-Normalized Event Surfaces (LNES), which preserves relative temporal information and polarity information within a temporal window.
Jiang~\etal~\cite{evhandpose} also used the LNES representation. Additionally, they proposed motion representations using shape flow and edge, effectively reducing motion ambiguity. They further addressed the issue of sparse annotations through a weakly-supervised learning framework.
Millerdurai~\etal~\cite{ev2hands} addressed the reconstruction of both hands using a point cloud-based approach.
Using a point cloud representation called Event Cloud, they preserved temporal information and effectively leveraged the raw event stream.

All of the studies mentioned above are limited to fixed third-person camera situations.
While there are many studies using conventional cameras  from an egocentric viewpoint~\cite{everyday_ego, AssemblyHands, EgoRGBD}, there are no studies using event cameras.
In egocentric event cameras, the task becomes more difficult because a large number of events are generated by the movements of the camera wearer.
In this work, we address this challenge by extracting and utilizing hand regions from the abundant events generated in egocentric scenarios.

\begin{figure*}[t]
\centerline{\includegraphics[width=\textwidth]{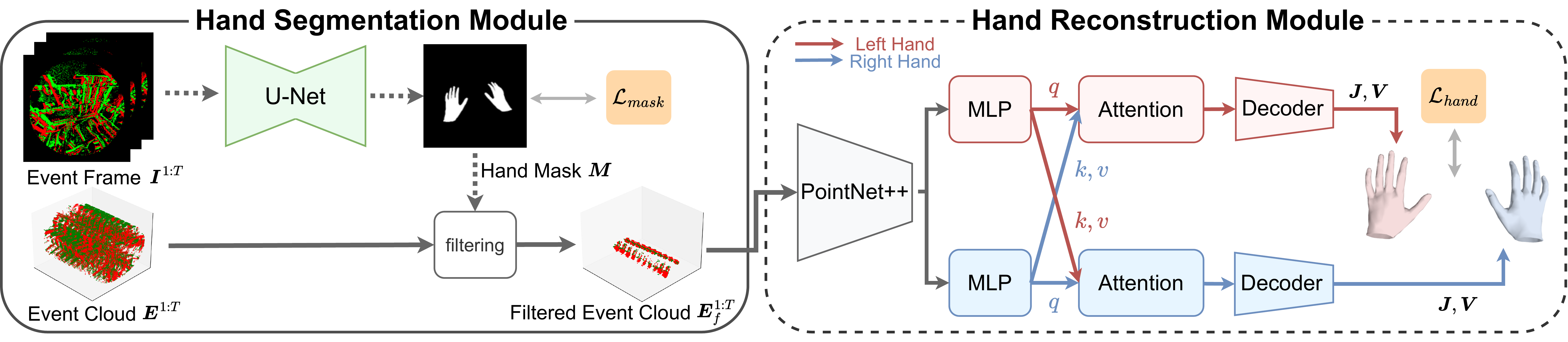}}
\caption{\textbf{The overview of \method.} 
}
\label{fig:model}
\end{figure*}


\section{Proposed Method}
\label{sec:method}

We propose \method, a 3D hand mesh reconstruction method that uses only event data in dynamic scenes from an egocentric view.
As shown in \cref{fig:model}, given a sequence of event frame $\bm{I}^{1:T}$ and event point clouds $\bm{E}^{1:T}$ segmented into $T$ fixed-width time windows, \method reconstructs 3D joint positions $\bm{J} \in \mathbb{R}^{20 \times 3}$ and mesh vertex positions $\bm{V} \in \mathbb{R}^{778 \times 3}$ of both human hands.
Our approach consists of two key components: 1) the Hand Segmentation Module, which extracts events triggered around the hand, and 2) the Hand Reconstruction Module, which outputs hand mesh. These are described in \cref{ssec:segmodule} and \cref{ssec:handmodule}, respectively, followed by an explanation of the loss function in \cref{ssec:loss}.

\subsection{Hand Segmentation Module}
\label{ssec:segmodule}
The Hand Segmentation Module takes an event frame $\bm{I}^{1:T} \in \mathbb{R}^{260 \times 346 \times 2}$ as input and predicts a hand region mask $\bm{M} \in \mathbb{R}^{260 \times 346 \times 1}$ to extract only events in the hand region from the event point cloud representation $\bm{E}^{1:T} \in \mathbb{R}^{N \times 5}$, where $N$ is the number of points. 
Here, we applied LNES~\cite{EventHands} to generate $\bm{I}^{1:T}$ from the raw events, since the LNES is known for preserving temporal information by applying temporal weights. We also applied Event Cloud~\cite{ev2hands} to generate $\bm{E}^{1:T}$ from the raw events, since it further preserves the raw temporal information.
We also used U-Net~\cite{unet} as the backbone network of the mask generation. 
The module takes multiple time-step frames $\bm{I}^{1:T}$ as input and outputs a single mask $\bm{M}$ for the latest time step, allowing it to leverage previous information to manage situations where fewer events occur.
The mask $\bm{M}$ is used to extract events from the Event Cloud $\bm{E}^{1:T}$ that correspond to the mask pixels. The filtered Event Cloud $\bm{E}_{f}^{1:T} \in \mathbb{R}^{N_f \times 5}$ is utilized as the input to the Hand Reconstruction Module.

This segmentation not only improves the accuracy of hand mesh reconstruction by effectively extracting the hand region but also contributes to a lower computational cost by minimizing the data to be processed.

\vspace{-1mm}

\subsection{Hand Reconstruction Module}
\label{ssec:handmodule}
The Hand Reconstruction Module takes a filtered Event Cloud $\bm{E}_{f}^{1:T} \in \mathbb{R}^{N_f \times 5}$ from the previous module and estimates the 3D joint positions $\bm{J} \in \mathbb{R}^{20 \times 3}$ and mesh vertex positions $\bm{V} \in \mathbb{R}^{778 \times 3}$ of both human hands.
The architecture of this module is inspired by Ev2Hands~\cite{ev2hands}.
PointNet++~\cite{pointnet2} is used as the backbone to extract features from the filtered Event Cloud $\bm{E}_{f}^{1:T}$.
Unlike Ev2Hands, there are no labels for the left hand, right hand, or background for each event, so instead of using a classifier for each event, we used Cross-Attention to learn the relationship between both hands.
In Cross-Attention, the query is taken from one branch, while the key and value are taken from the opposite branch.
Finally, the decoder produces 3D hand joint positions $\bm{J} \in \mathbb{R}^{20 \times 3}$ and 3D hand mesh vertices $\bm{V} \in \mathbb{R}^{778 \times 3}$ as output using the MANO~\cite{MANO} model, a parametric representation of the human hand mesh.

MANO parameters include a pose vector $\boldsymbol{\theta} \in \mathbb{R}^{15}$ and a shape vector $\boldsymbol{\beta} \in \mathbb{R}^{10}$.
Additionally, we encoded the rigid transformation parameters, including translation $\boldsymbol{t} \in \mathbb{R}^3$ and rotation $\boldsymbol{R} \in \mathbb{R}^3$, for each hand.
From the MANO model, we can extract sparse hand joints and hand mesh vertices for each hand individually using the function $\boldsymbol{J}, \boldsymbol{V} = \mathcal{J}(\boldsymbol{\theta}, \boldsymbol{\beta}, \boldsymbol{t}, \boldsymbol{R})$.
The resulting joint locations $\boldsymbol{J} \in \mathbb{R}^{20 \times 3}$ represents the 3D positions of the regressed hand joints, while the mesh vertices $\boldsymbol{V} \in \mathbb{R}^{778 \times 3}$ correspond to the 3D coordinates of the hand mesh. 
For simplicity, we used the same notation for the hand parameters for both the left and right hands, unless explicitly specified otherwise.

\vspace{-1mm}

\subsection{Loss Function}
\label{ssec:loss}

\noindent\textbf{Mask Loss.}
This paper adopts a combination of Binary Cross Entropy (BCE) Loss and Dice Loss~\cite{dice_loss} for segmentation tasks.
These loss functions complement each other, where BCE Loss evaluates the probabilistic error for each pixel, and Dice Loss measures the overlap between the predicted and ground-truth segmentation.
This combination enhances the accuracy of mask prediction.
The BCE Loss, Dice Loss, and total mask loss are defined as follows:
\begin{equation}
\mathcal{L}_{\text{BCE}} = - \frac{1}{N} \sum_{i=1}^N \Big[ y_i \log(\hat{y}_i) + (1 - y_i) \log(1 - \hat{y}_i) \Big],
\end{equation}

\begin{equation}
\mathcal{L}_{\text{Dice}} = 1 - \frac{2 \sum_{i=1}^N y_i \hat{y}_i}{\sum_{i=1}^N y_i + \sum_{i=1}^N \hat{y}_i},
\end{equation}

\begin{equation}
\mathcal{L}_{\text{mask}} = \lambda_{\alpha} \mathcal{L}_{\text{BCE}} + \lambda_{\beta} \mathcal{L}_{\text{Dice}},
\end{equation}
where $N$ is the number of pixels, $y_i \in \{0, 1\}$ is the ground-truth mask label, and $\hat{y}_i \in [0, 1]$ is the predicted probability.
The $\lambda_{\alpha}$ and $\lambda_{\beta}$ are weights that balance the contributions of the BCE Loss and Dice Loss, respectively.

\noindent\textbf{3D Hand Joints Loss.}
The loss functions for evaluating the accuracy of 3D hand joint estimations are defined as follows.
The 3D hand joints loss $\mathcal{L_{\text{joints}}}$ is defined as the L1 distance between the predicted and ground-truth joint positions, while the interaction hand joints loss $\mathcal{L_{\text{interhand}}}$ is defined as the
L2 distance between the left and right hands to measure their positional consistency:
\begin{equation}
    \mathcal{L_{\text{joints}}} = \frac{1}{N_J}\sum_{i=1}^{N_J} \| \bm{\hat{J}_i} - \bm{J_i} \|_1,
\end{equation}

\begin{equation}
    \mathcal{L_{\text{interhand}}} = \frac{1}{N_J} \sum_{i=1}^{N_J} \| (\bm{\hat{J}}_{\text{left}, i}-\bm{\hat{J}}_{\text{right}, i})-(\bm{J}_{\text{left}, i}-\bm{J}_{\text{right}, i}) \|_2,
\end{equation}
where \( \bm{\hat{J}}_i \) and \( \bm{J}_i \) are the predicted and ground-truth \( i \)-th 3D hand joints, and \( N_J \) is the number of hand joints.

\noindent\textbf{3D Hand Mesh Vertices Loss.}
The loss functions for evaluating the accuracy of 3D hand mesh vertex estimations are defined as follows.
The 3D hand mesh vertices loss $\mathcal{L_{\text{vertices}}}$ is defined as the L1 distance between the predicted and ground-truth vertex positions, ensuring the correctness of the estimated hand mesh structure:
\begin{equation}
    \mathcal{L_{\text{vertices}}} = \frac{1}{N_V}\sum_{i=1}^{N_V} \| \bm{\hat{V}_i} - \bm{V_i} \|_1,
\end{equation}
where \( \bm{\hat{V}}_i \) and \( \boldsymbol{V}_i \) are the predicted and ground-truth \( i \)-th 3D hand mesh vertex positions, and \( N_V \) is the number of vertices.

\noindent\textbf{MANO Loss.} 
The MANO loss measures the difference between the predicted and ground-truth MANO pose parameters $\bm{\theta}$ and shape parameters $\bm{\beta}$.
This loss encourages the model to generate accurate hand pose and shape representations:
\begin{equation}
    \mathcal{L_{\text{MANO}}} = \| \hat{\bm{\theta}} - \bm{\theta} \|_2 + \| \hat{\bm{\beta}} - \bm{\beta} \|_2.
\end{equation}

\noindent\textbf{Total Hand Loss.} 
The total hand loss $\mathcal{L_{\text{hand}}}$ combines all the individual loss components to optimize the model, and is as follows:
\begin{align}
\mathcal{L_{\text{hand}}} = \lambda_{\gamma}\mathcal{L_{\text{joints}}} 
+ \lambda_{\delta}\mathcal{L_{\text{interhand}}}
+ \lambda_{\epsilon}\mathcal{L_{\text{vertices}}}  
+\lambda_{\zeta}\mathcal{L_{\text{MANO}}},
\end{align}
where the weights $\lambda_{\gamma}, \lambda_{\delta}, \lambda_{\epsilon}$, and $\lambda_{\zeta}$ correspond to $\mathcal{L_{\text{joints}}}$, $\mathcal{L_{\text{interhand}}}$, $\mathcal{L_{\text{vertices}}}$, and $\mathcal{L_{\text{MANO}}}$, respectively.

\begin{figure}[t]
\centerline{\includegraphics[width=\linewidth]{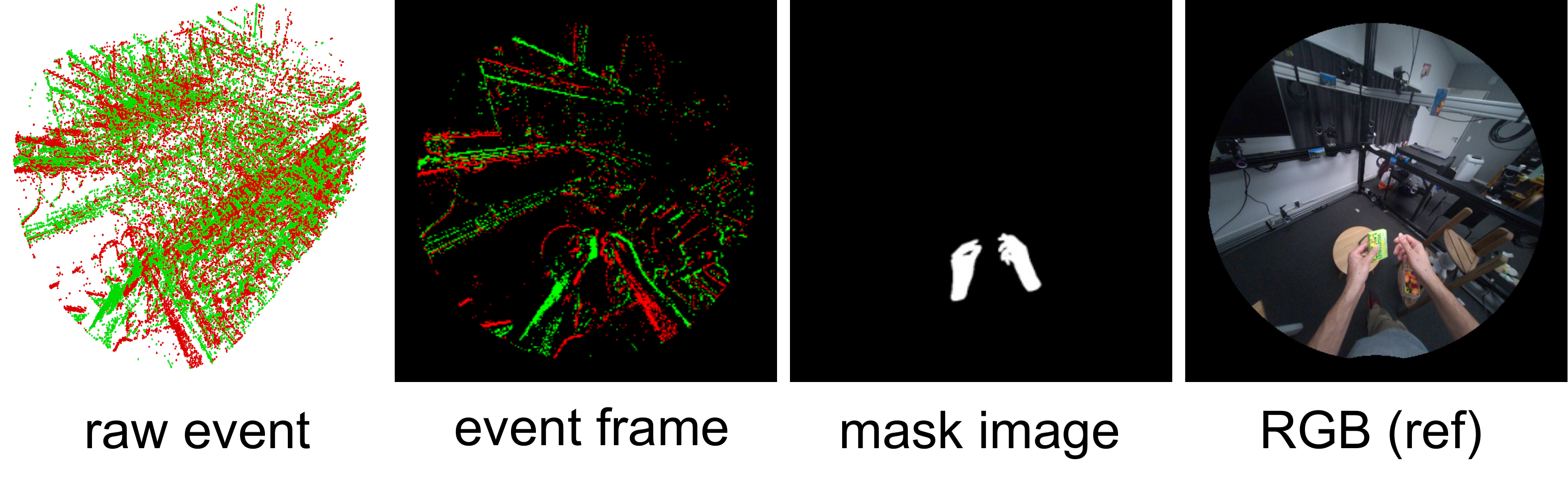}}
\caption{\textbf{Sample from \dataset.} We used the MANO ground-truth annotations and RGB image directly from HOT3D~\cite{hot3d}, while independently providing the raw event and hand mask.} 
\label{fig:dataset}
\end{figure}

\begin{figure*}[t]
\centerline{\includegraphics[width=\textwidth]{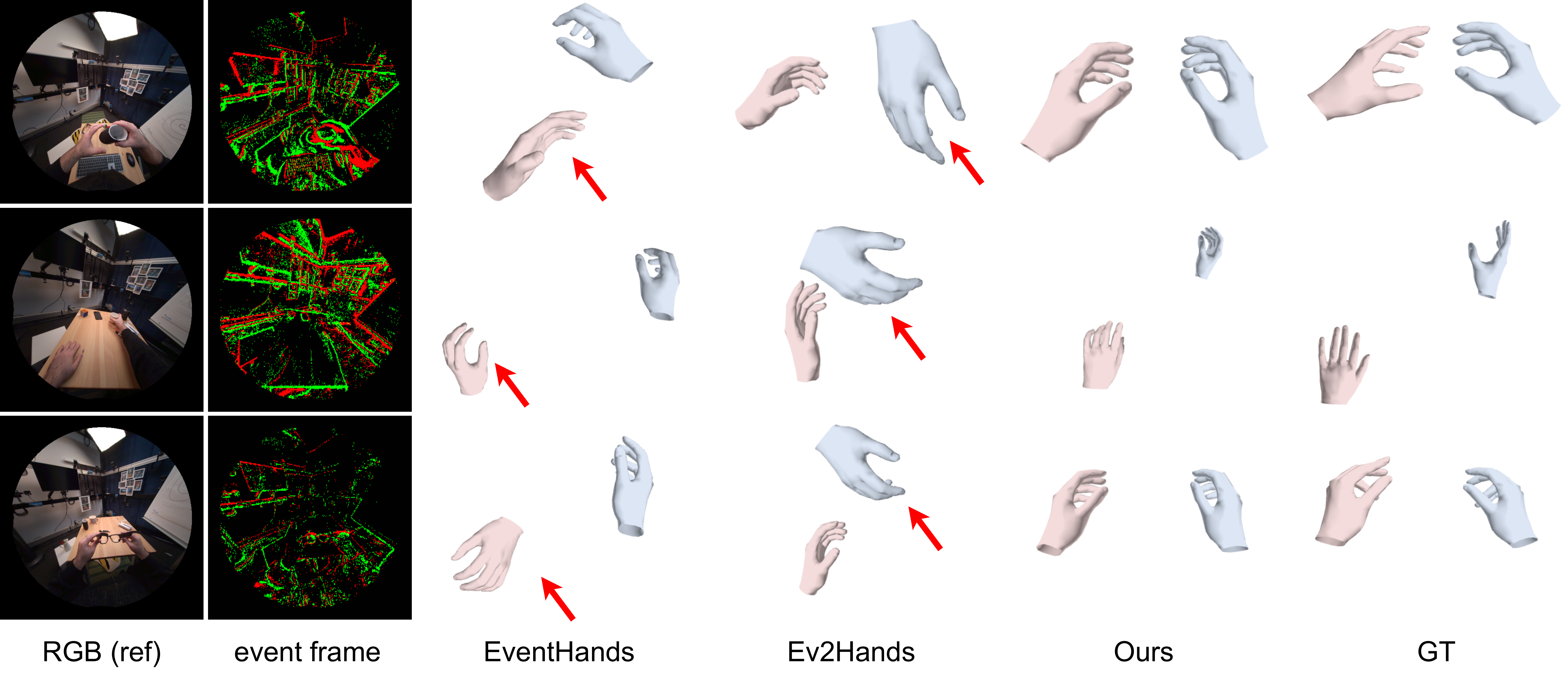}}
\caption{\textbf{Qualitative Evaluation.}
We compared our method with EventHands~\cite{EventHands} and Ev2Hands~\cite{ev2hands} as baselines. Red arrows indicate the failure parts. RGB images were not used as input, but only as a reference.}
\label{fig:result}
\end{figure*}


\section{EXPERIMENT}
\label{sec:exp}

\subsection{Dataset}
\label{ssec:dataset}
To train our model, we needed event data with ground-truth annotations of both hands from an egocentric view.
However, there are no datasets from an egocentric view. 
Thus, we created an event dataset \dataset using the event simulator v2e~\cite{v2e} from the HOT3D~\cite{hot3d} dataset.
We specifically used a subset of the Aria glasses data within the HOT3D.
Our dataset included a total of $447{,}704$ samples, which are divided into $347{,}854$ samples for training and $99{,}850$ samples for evaluation.
Sample data from \dataset are illustrated in \cref{fig:dataset}.
We used MANO parameters, camera extrinsic parameters, and intrinsic parameters given by HOT3D.
Using these values, we conducted distortion correction on the videos and then converted the corrected videos using the event simulator.
The output events size was set to $346 \times 260$, based on the resolution of the DAVIS346~\cite{davis346} event camera, which is also used in Ev2Hands~\cite{ev2hands}.
Additionally, we projected the provided 3D mesh ground-truth annotations onto the 2D space to create ground-truth hand masks.

\subsection{Implementation Details}
\label{ssec:implement}
We used all \dataset data~(447K) for the Hand Segmentation Module and the Hand Reconstruction Module, splitting it into approximately 278K~(62\%) for training, 70K~(16\%) for validation, and 99K~(22\%) for testing.
We trained the Hand Segmentation Module using a single NVIDIA RTX 6000 Ada Generation GPU with a batch size of 64 and 10 epochs. We used Adam~\cite{adam} Optimizer with a learning rate of $1 \cdot 10^{-5}$.
We set $\lambda_{\alpha}=0.7, \lambda_{\beta}=0.3$.
We train the Hand Reconstruction Module using the same GPU with a batch size of 16 and 10 epochs. We used Adam Optimizer with a learning rate of $1 \cdot 10^{-5}$.
We set $\lambda_{\gamma} = 1 \cdot 10^{-1},\lambda_{\delta}=1.0, \lambda_{\epsilon}=1.0, \lambda_{\zeta} = 20$.
The number of fixed-width time windows $T$ is $3$.

\subsection{Baseline Methods}
\label{ssec:compare}
\noindent\textbf{EventHands}~\cite{EventHands} is a frame-based method that uses only event data as input. 
Because it is recognized as one of the state-of-the-art methods from a third-person view, we used it in our experiments.
This method outputs predictions only one hand at a time. Therefore, we trained separate models for the left and right hands using \dataset.

\noindent\textbf{Ev2Hands}~\cite{ev2hands} is a point cloud-based approach that also uses only event data.
It is the only prior work that outputs both hands.
This method requires event data labeled as left hand, right hand, or background, but \dataset does not provide these labels.  
Therefore, we first trained on the annotated Ev2Hands-S~\cite{ev2hands} dataset and then fine-tuned it with \dataset.

\subsection{Evaluation Metrics}
\label{ssec:eval}
Following Ev2Hands~\cite{ev2hands}, we used the Percentage of Correct Keypoints (PCK) and the area under the PCK curve (AUC) with thresholds from 0 to 100 millimeters (mm).
We utilized the Relative AUC (R-AUC) to evaluate the performance of 3D hand pose estimation. 
The R-AUC is based on joint positions relative to the wrist joint of each hand to measure the accuracy of 3D joints positions for each hand independently.
In addition, we used the Mean Per Joint Position Error (MPJPE) and Mean Per Vertex Position Error (MPVPE) in millimeters, which are commonly used metrics for evaluating 3D hand mesh reconstruction tasks.

\section{RESULTS}
\label{sec:result}

\subsection{Quantitative Evaluation}
\label{ssec:quant}
To evaluate the effectiveness of our proposed method, we conducted a quantitative evaluation with baseline methods; the results are shown in \Cref{table:quant}. The results demonstrate that the proposed method outperforms all baselines.
In particular, our method improved the R-AUC by approximately 0.2 points compared to the baseline. Additionally, MPJPE improved by about $45~\mathrm{mm}$ and MPVPE by about $20~\mathrm{mm}$.

To further analyze the effectiveness of the Hand Segmentation Module and Cross-Attention, we conducted an ablation study.
 We evaluated the performance without the Hand Segmentation Module and with Self-Attention in the Hand Reconstruction Module, instead of Cross-Attention. 
In Self-Attention, the query, key, and value were all taken from the same branch.
The results show that each of our proposed components worked effectively.

\subsection{Qualitative Evaluation}
\label{ssec:quali}
We conducted a qualitative evaluation to verify the accuracy of hand reconstruction; the results are shown in \cref{fig:result}. 
The results demonstrate that our proposed method is the closest to the ground truth.
Our proposed method reconstructs the hand mesh accurately, whereas the baseline methods fail due to ambiguities introduced by numerous background events, resulting in errors in hand orientation and the distance between the two hands.
However, even with our proposed method, it is difficult to represent the detailed fingertip movements of the ground truth.

\begin{table}[t]
    \centering
    \caption{\textbf{Quantitative Evaluation.} Ablation studies include evaluations without segmentation and with Self-Attention to analyze the contributions of the components. The best values are shown in \textbf{bold}. }\medskip
    \label{table:quant}
    \scalebox{0.75}{
    \begin{tabular}{lccc}
        \toprule
          & R-AUC ($\uparrow$) & MPJPE [mm] ($\downarrow$) & MPVPE [mm] ($\downarrow$) \\
        \midrule
        EventHands~\cite{EventHands}  & 0.243 & 105.80 & 65.23 \\
        Ev2Hands~\cite{ev2hands} & 0.251 & 106.75 & 69.66 \\
        \cmidrule(l){1-4} 
        Ours w/o Segmentation & 0.392 & 66.24 & 47.11  \\
        Ours w/ Self-Attention & 0.431 & 62.84 & 44.95  \\
        \cmidrule(l){1-4} 
        \cmidrule(l){1-4} 
        \rowcolor{lightgray}
        Ours & \textbf{0.450} & \textbf{59.51} & \textbf{42.92}  \\
        \bottomrule
    \end{tabular}}
\end{table}


\section{Conclusion}
\label{sec:conc}
In this study, we proposed \method, the first method of event-based egocentric 3D hand mesh reconstruction.
Our Hand Segmentation Module extracts only events in the hand region, thereby reducing the influence of background events caused by the motion of the camera wearer and reducing computational cost by minimizing the event data to be processed.
As many other hand mesh reconstruction methods, our method often fails when the hand is occluded by an object. For future work, hand-object interaction should be considered to make the method more robust. We hope that our \method will contribute to further developments in this research field.\\

\noindent
{\textbf{{Acknowledgment.}}
\noindent
This work was partially supported by JST Presto JPMJPR22C1 and Keio University Academic Development Funds. Masashi Hatano was supported by JST BOOST, Japan Grant Number JPMJBS2409.

\vfill
\pagebreak

\bibliographystyle{IEEEbib}
\bibliography{refs}

\end{document}


%

\maketitle

\hypersetup{linkcolor=black}
\tableofcontents
\hypersetup{linkcolor=red}

\section{Overview of the Supplementary Material}
This supplementary document contains additional details and discussions of our \method. 
Please also refer to the \href{https://sigport.org/documents/supplementary-material-eventegohands-event-based-egocentric-3d-hand-mesh-reconstruction}{supplementary video} for video qualitative evaluation.
We highlight reference numbers assosiated with the main paper in \textcolor{blue}{blue}, and those associated with this supplementary document in \textcolor{red}{red}.

\section{Hand Segmentation Module}
\label{sec:maskresult}

\begin{figure}[t]
\centerline{\includegraphics[width=\linewidth]{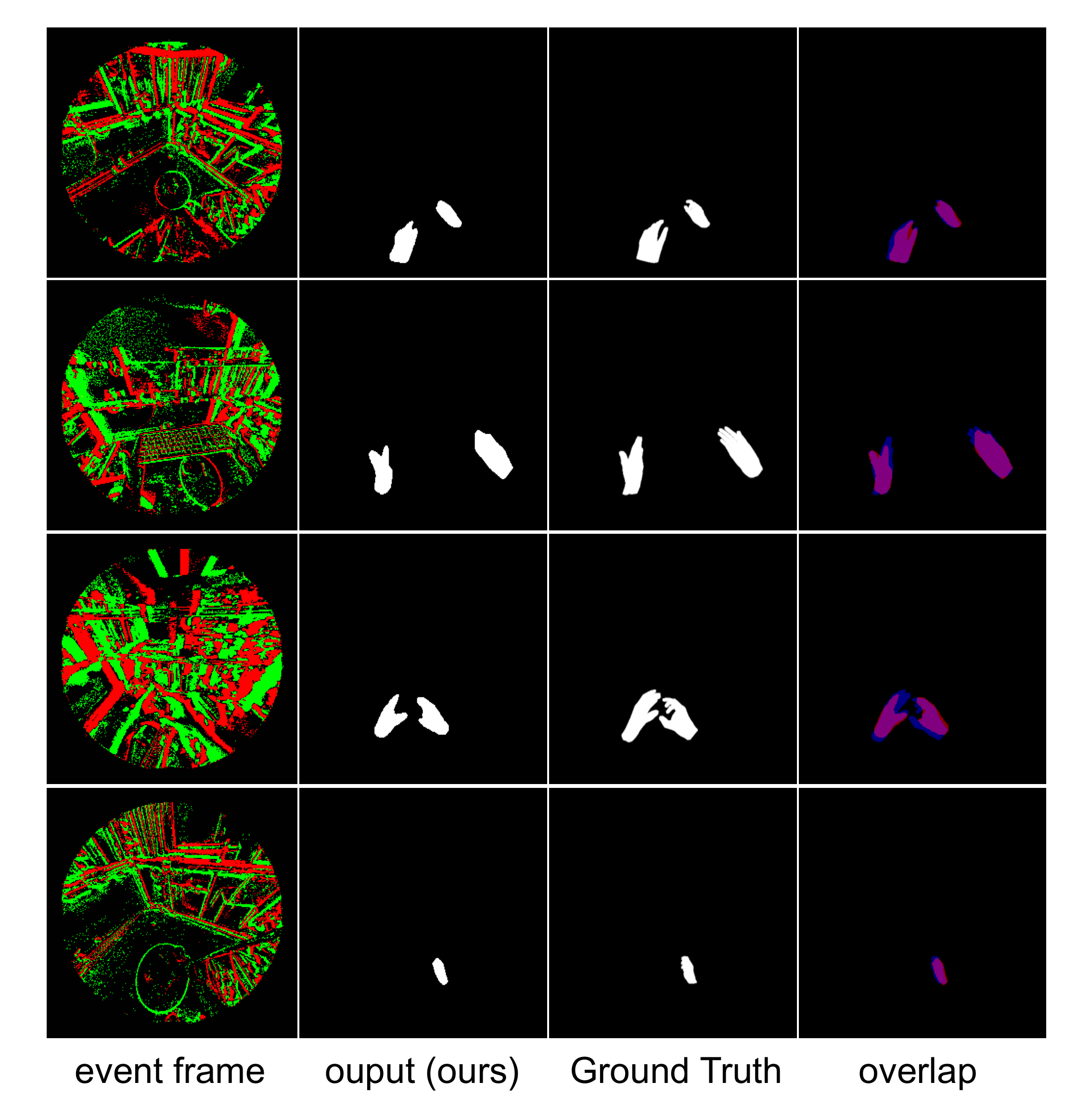}}
\caption{
\textbf{Hand segmentation results.} We show qualitative results of our hand segmentation. In the overlap column, the mask in \textcolor{red}{red} and \textcolor{blue}{blue} represents our predicted hand mask and ground truth, respectively. The overlap region between two is shown in \textcolor{purple}{purple}.
}
\label{fig:maskres}
\end{figure}

\subsection{Locally-Normalised Event Surfaces (LNES)}
\label{ssec:lnes}
Locally-Normalised Event Surfaces (LNES)~\cite{EventHands} is one of the frame-based event representations.
The LNES preserves temporal information within a time window by applying temporal weights.
The LNES representation $\bm{I}$ is expressed by the following equation:
\begin{equation}
\bm{I}(x_i, y_i, p_i) = \frac{t_i - t_{s}}{t_l - t_{s}},
\end{equation}
where $x_i, y_i$ are the pixel coordinates, $p_i$ represents the polarity, and $t_i$ is the timestamp of the \( i \)-th event point. Here, $t_{s}$ is the timestamp of the first event, and  $t_{l}$ is the timestamp of the last event within a time window.

We adopt LNES as the event-frame representation for the Hand Segmentation Module.
Our method takes $\bm{I}^{1:T}$ as input, which is a continuous LNES $\bm{I}$ within a fixed-width time window $T$.
When the camera wearer’s movement is slight, the number of triggered events decreases, which can result in a decline in hand estimation accuracy. We can address situations where fewer events occur by incorporating multiple time steps as input. 
In our proposed method, $T = 3$ is used, which corresponds to a very short interval of 3 frames at 30 fps. 
Since the mask does not change significantly within a short time interval, the mask at the latest timestamp is used as the filtering mask.

\subsection{Event Cloud}
\label{ssec:cloud}
The Event Cloud~\cite{ev2hands} is one of the point-cloud-based event representations.
A single event point is represented as follows:
\begin{equation}
    \bm{E_k} = (x_k, y_k, t_k, p_k, n_k),
\end{equation}
where $x_k, y_k$ are the pixel coordinates, $t_k$ is the timestamp, and $p_k, n_k$ are the positive and negative polarity of the \(k\)-th event point.
The Event Cloud $\bm{E} \in \mathbb{R}^{N \times 5}$ is a point cloud consisting of $N$ event points $\bm{E_k}$. The total number of event points is $N = 2048$.

\subsection{Qualitative Results}
\label{ssec:res}
The results of the mask region estimated by the Hand Segmentation Module are shown in \cref{fig:maskres}.
The results indicate that the Hand Segmentation Module can estimate the hand region from LNES even when only one hand is visible.


\begin{figure}[t]
  \begin{subfigure}{0.24\textwidth}
    \includegraphics[width=\linewidth]{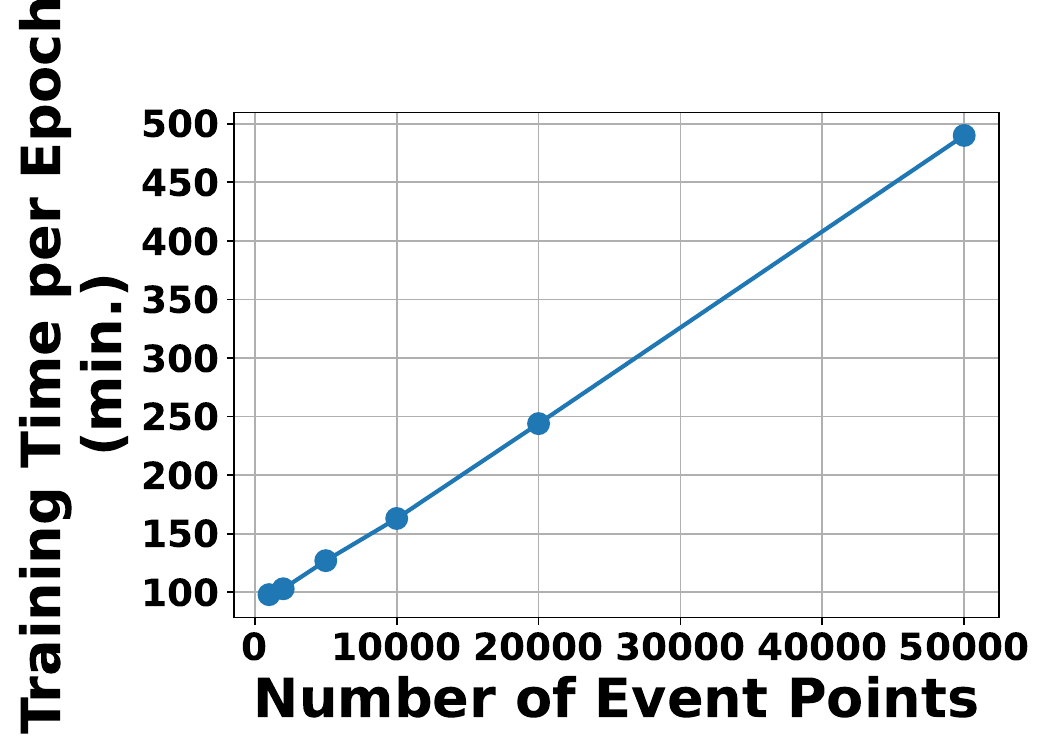}
    \caption{Training time cost}
    \label{fig:figure1}
  \end{subfigure}%
  \hfill
  \begin{subfigure}{0.24\textwidth}
    \includegraphics[width=\linewidth]{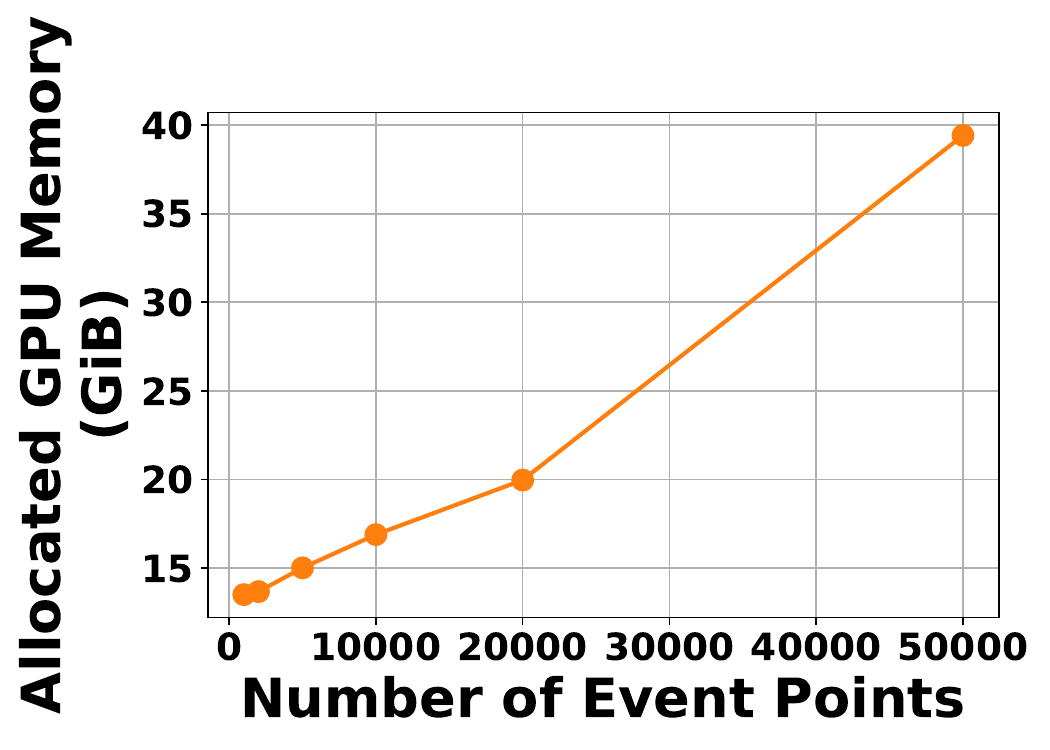}
    \caption{GPU memory consumption}
    \label{fig:figure2}
  \end{subfigure}
  \caption{\textbf{Analysis of computational cost w.r.t the number of event points}.}
  \label{fig:cost_graph}
\end{figure}

\section{Analysis of Computational Cost}
\label{sec:cost}
To investigate how the Hand Segmentation Module contributes to reducing computational cost by decreasing the number of events, we conducted an analysis of computational cost.
We analyze the training time per epoch (in minutes) and GPU memory usage (in gibibytes) while varying the number of event points.
Most of the \dataset datasets we created contain between a few thousand (K) and $100{,}000$ (100K) event points within a single time window at 30 fps.
The number of events in our proposed method is $N = 2048$.
Therefore, we can reduce the number of events by approximately 1/50 at most.
As shown in \cref{fig:cost_graph}, training time and GPU memory usage increase as the number of events increases. Therefore, reducing the number of events helps to lower the computational cost.

\section{Video Qualitative Evaluation}
\label{sec:video}
We include the \href{https://sigport.org/documents/supplementary-material-eventegohands-event-based-egocentric-3d-hand-mesh-reconstruction}{supplementary video} showcasing the 3D hand mesh reconstruction performance of \method.
The video is also available on the page where this document can be found.
The video contains the qualitative evaluation results, including comparisons with the baseline methods and the ablation study.
The video demonstrates that our method produces the most stable and closest output to the ground truth.
The \method is designed for single time window output without considerations for temporal
coherence, which may result in jittery outputs when applied to video evaluation.

\begin{table}[t]
    \centering
    \caption{\textbf{Balancing Hyperparameters for Hand Segmentation Module}. The yellow row indicates the value we adopted.}
    \label{tab:mask_loss}
    \begin{tabular}{lc}
        \toprule
         $\lambda_{\alpha}:\lambda_{\beta}$ & IoU (↑) \\
        \midrule
        0.8 : 0.2 & 0.529 \\
        \rowcolor{yellow} 0.7 : 0.3 & \textbf{0.567} \\
        0.6 : 0.4 & 0.519 \\
        \bottomrule
    \end{tabular}
\end{table}

\begin{table}[t]
    \centering
    \caption{\textbf{Balancing Hyperparameters for Hand Reconstruction Module}. The yellow row indicates the value we adopted.}
    \label{tab:hand_loss}
    \scalebox{0.8}{
    \begin{tabular}{lccc}
        \toprule
        $\lambda$ & R-AUC ($\uparrow$) & MPJPE [mm] ($\downarrow$) & MPVPE [mm] ($\downarrow$) \\
        \midrule
        $\lambda_{\gamma} = 1$ & 0.440 & 60.89 & 43.78 \\
        \rowcolor{yellow} $\lambda_{\gamma} = 0.1$  & \textbf{0.450} & \textbf{59.51} & \textbf{42.92} \\
        $\lambda_{\gamma} = 0.01$ & 0.419 & 63.26 & 44.71 \\
        \midrule
        $\lambda_{\delta} = 10$   & 0.440 & 60.97  & 43.83 \\
        \rowcolor{yellow} $\lambda_{\delta} = 1$    & \textbf{0.450} & \textbf{59.51} & \textbf{42.92} \\
        $\lambda_{\delta} = 0.1$  & 0.446 & 59.94  & 43.16 \\
        \midrule
        $\lambda_{\epsilon} = 10$   & 0.444 & 60.24 & 43.53 \\
        \rowcolor{yellow} $\lambda_{\epsilon} = 1$  & \textbf{0.450} & \textbf{59.51} & \textbf{42.92} \\
        $\lambda_{\epsilon} = 0.1$  & 0.441 & 61.12 &  43.98 \\
        \midrule
        $\lambda_{\zeta} = 30$   & 0.383 & 67.92 & 47.68 \\
        \rowcolor{yellow} $\lambda_{\zeta} = 20$    & \textbf{0.450} & \textbf{59.51} & \textbf{42.92} \\
        $\lambda_{\zeta} = 10$  & 0.441 & 60.93 & 43.60 \\
        \bottomrule
    \end{tabular}
    }
\end{table}

\section{Ablation Analysis}
\label{sec:ablation}

We conduct an ablation study to analyze the impact of balancing hyperparameters for losses in Hand Segmentation Module and Hand Reconstruction Module.
We use predefined sets of loss weights for Hand Segmentation Module: $(\lambda_{\alpha}, \lambda_{\beta})\in\{(0.8,0.2),(0.7,0.3),(0.6,0.4)\}$.
Regarding Hand Reconstruction Module, we systematically vary the values of each loss weight using the following ranges: $\lambda_{\gamma}$ from 0.01 to 1, $\lambda_{\delta}$ from 0.1 to 10, $\lambda_{\epsilon}$ from 0.1 to 10, and $\lambda_{\zeta}$ from 10 to 30.
We vary one loss weight at a time while keeping the others fixed at their default value: 0.1, 1, 1, and 20 for $\lambda_{\gamma}$, $\lambda_{\delta}$, $\lambda_{\epsilon}$, and $\lambda_{\zeta}$, respectively.
We use the Intersection over Union (IoU) as a metric for hand segmentation, which is commonly employed in segmentation tasks, to assess the overlap between the predicted mask and the ground truth mask.
R-AUC, MPJPE, and MPVPE are adopted as metrics of hand reconstruction.
\cref{tab:mask_loss} and \cref{tab:hand_loss} show comparisons of hyperparameters for the Hand Segmentation and Hand Reconstruction Module, respectively.

\begin{figure}[t]
\centerline{\includegraphics[width=\linewidth]{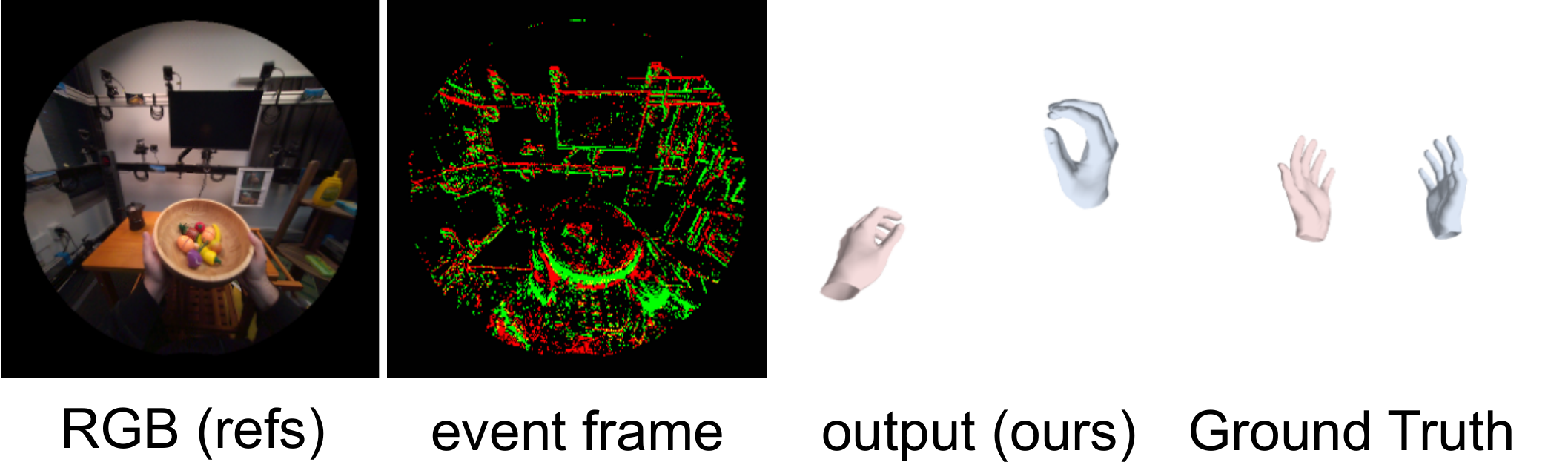}}
\caption{
\textbf{Example of failure case.}
}
\label{fig:fail}
\end{figure}

\section{Failure Case}
\label{sec:fail}
Although the proposed method demonstrates improvements over existing methods, there is a challenge when interacting with objects.
\cref{fig:fail} illustrates a failure case where, like many other 3D hand mesh reconstruction methods, our method struggles when an object occludes the hand. Since our method does not consider hand-object interactions, it will be necessary to incorporate this aspect in future work.


\bibliographystyle{IEEEbib}
\bibliography{supp_refs}